# Auto-Encoder-Extreme Learning Machine Model for Boiler NOx Emission Concentration Prediction

*Zhenhao Tang[a,*], Shikui Wang[a], Xiangying Chai[a], Shengxian Cao[a], Tinghui Ouyang[b], Yang Li[c]*

[a] *College of Automation Engineering, Northeast Electric Power University, Jilin, China*
[b] *Artificial Intelligence Research Center (AIRC), National Institute of Advanced Industrial Science and Technology (AIST), Tokyo, Japan*
[c] *School of Electrical Engineering, Northeast Electric Power University, Jilin, China*

**Abstract**

An automatic encoder (AE) extreme learning machine (ELM)-AE-ELM model is proposed to predict the NOx emission concentration based on the combination of mutual information algorithm (MI), AE, and ELM. First, the importance of practical variables is computed by the MI algorithm, and the mechanism is analyzed to determine the variables related to the NOx emission concentration. Then, the time delay correlations between the selected variables and NOx emission concentration are further analyzed to reconstruct the modeling data. Subsequently, the AE is applied to extract hidden features within the input variables. Finally, an ELM algorithm establishes the relationship between the NOx emission concentration and deep features. The experimental results on practical data indicate that the proposed model shows promising performance compared to state-of-art models.

[*] Corresponding author.
   Email address: tangzhenhao@neepu.edu.cn (Z. Tang).

## 1. Introduction

The energy supply structure has been recently optimized and adjusted in China, with renewable electricity installed capacity reaching 39.5% of the total power installed capacity in 2019 [1]. Regardless of those renewable energy sources such as wind power and photovoltaic, coal-fired power plants still contributed 72% of the total electricity in 2018. That is, the traditional thermal power plants still occupy the dominant position in electricity production in China [2]. Meanwhile, nitrogen oxides (NOx) of coal-fired boilers in power plants cause environmental pollution and endanger human health. In order to limit the NOx emission concentration, the national emissions standard for NOx emission concentration of coal-fired power plants was reduced to 50mg/m$^3$ in 2019 in China [3]. Currently, coal-fired power plants adopt combustion optimization [4–6]and flue gas denitration technology [7] to reduce NOx emission concentration. Controlling boiler manipulated parameters such as secondary air and coal feed amount can improve the combustion state of the boiler. For this purpose, artificial intelligence technologies and optimization algorithms can determine the precise combination of manipulated parameters. This combustion optimization reduces NOx emission concentration while improving thermal efficiency. However, the distributed control system (DCS) has the characteristics of measurement lag, electromagnetic interference, and high maintenance costs, resulting in the inability to obtain the correspondence between NOx emission concentration and operating parameters in real-time. Therefore, establishing an accurate real-time NOx prediction model is the basis of combustion optimization [8].

The NOx estimation methods can mainly be classified into mechanism-based, computational fluent dynamic (CFD) simulation, and data-driven methods. In terms of mechanism modeling, Zhao et al. [9] established a mathematical model for describing NOx concentration based on the combustion mechanism. Jones et al. [10] studied the influence factors of NOx emission concentration through tube drop furnace experiments and empirical formula. Despite its real-time prediction of the NOx concentration, the method assumed many ideal conditions. In addition, the online prediction accuracy with mechanism-based methods was not stable due to the complexity and instability of combustion. Subsequently, CFD simulation method was explored in predicting NOx emission concentration. Schluckner et al. [11] proposed a CFB-based prediction model for NOx emission concentration. Chang et al. [12] proposed a CFD model including coal combustion and NOx formulation. The CFD-based methods could obtain the distribution of NOx in the boiler and have played a great role in boiler design. However, the CFD-based methods are time-consuming and not robust to combustion condition change.

The data-driven modeling methods have been widely exploited to improve prediction accuracy by utilizing production data from the DCS. Zhou et al. [13] applied support vector regression (SVR) and ant colony algorithm (ACO) to build a NOx emission concentration model for a 300 MW tangentially fired dry bottom boiler. However, the above algorithm could not mine the deep feature information in the high-dimensional production data due to its shallow structure. In order to overcome this limitation, deep learning methods were studied in NOx emission concentration modeling. A deep neural network with a modified early stopping algorithm and least square support vector machine was developed to predict NOx emission concentration [14]. Xie et al. [15] proposed a long short-term memory (LSTM) network-based prediction model to accurately predict the future sequence of NOx emission concentration in the next horizon. The mean absolute percentage errors of the model were within 3%. Fang et al. [16] established three deep belief neural (DBN)-based NOx emission concentration prediction models for coal-fired power plants with a data set covering various operating conditions. Although deep learning networks showed good performance, they suffered from the problem of high computational complexity.

In order to improve the computing efficiency of neural networks during modeling, the extreme learning machine [17–19] (ELM) algorithm was successfully applied to many aspects, such as face recognition [20], solar radiation prediction [21], silicon prediction in blast furnace [22], and boiler efficiency prediction [23]. Peng et al.[24] applied ELM to predict the NOx emission concentration, achieving a model whose mean absolute error (MAE) was only 0.92%. These studies proved that the ELM is suitable for modeling complex industrial processes. In addition, the ELM is faster in the training process than traditional neural network algorithms.

Based on the field experience and combustion reaction mechanism, the amount of NOx emission concentration can be regulated by adjusting the manipulated variables of the boiler [4], such as primary airflow, secondary airflow, and separated over-fire air

(SOFA) flow. However, due to the complexity and variability of the boiler combustion process, the amount of NOx emission concentration can be affected by many factors, such as unit load, main steam temperature, etc. Therefore, selecting appropriate features used for constructing a prediction model is challenging. The feature extraction is the first step in the NOx emission concentration prediction model, preventing information redundancy. Principal component analysis (PCA) has been widely used to extract input data features [25]. The partial least-square method [26] (PLS) was conducted for the correlation analysis of the NOx emission concentration model.However, the above feature extraction algorithms could not extract deep features from high-dimensional nonlinear data. In order to extract the deep features in nonlinear data to deal with complex problems, deep learning has attracted extensive attention from many practical applications. Recently, the auto-encoder [27] (AE) showed outstanding ability to learn deep features. The AE consists of two sub-networks: encoder and decoder, and the loss function is defined as the difference between input and reconstructed data. Due to the deep network structure of the auto-encoder, it can perform nonlinear mapping of high-dimensional input data and dig deep feature information. Since the reconstructed data contains important feature information, the AE algorithm can be used to reduce inputs while keeping meaningful deep information.

This paper proposes the AE-ELM model to predict the NOx emission concentration of 1000MW coal-fired boilers. First, the NOx generation mechanism was employed to select a database of alternative variables. Second, the influence of delay time on modeling accuracy was analyzed using a mutual information algorithm, considering the characteristics of large delays in the production process of the power station boiler. Then, the data were reconstructed based on the results. Third, the AE algorithm extracts deep features. Finally, the ELM algorithm estimates the nonlinear relationship between inputs and the NOx emission concentration. The proposed method is evaluated on practical data compared to MLR, BP, RBF, and SVR.

The contributions of this paper mainly include three aspects:

(a) A MI-based method for calculating delay time between NOx emission concentration and alternative variables is developed, which is used to reconstruct the model data.

(b) A new feature extraction method based on an auto-encoder is developed to extract deep features from high-dimensional production data.

(c) An ELM-based model for NOx emission concentration prediction is proposed.

The rest of this paper is organized as follows. Section 2 describes the adopted boiler in this study and the data preprocessing. In section 3, the prediction model is presented in detail. Section 4 gives the experimental results and discussion. Lastly, Section 5 concludes this paper.

## 2. The boiler and data preprocessing

### 2.1 Description of the studied boiler

A once-through boiler of 1000 MW ultra-supercritical coal-fired units of thermal power is studied in this paper. As illustrated in Fig. 1, the boiler has a large furnace with a 15,670 mm×15,670 mm cross-section and a 32,084 mm height, and low NOx axial swirl burners are installed on the front wall. There are six levels of secondary air burners (AA-EF) installed in the four separate corners of the cross-section. Primary and secondary air nozzles are set by the interval. One level of over-fire air exists above the primary combustion zone. The schematic diagram of the studied boiler is given in Fig. 1.

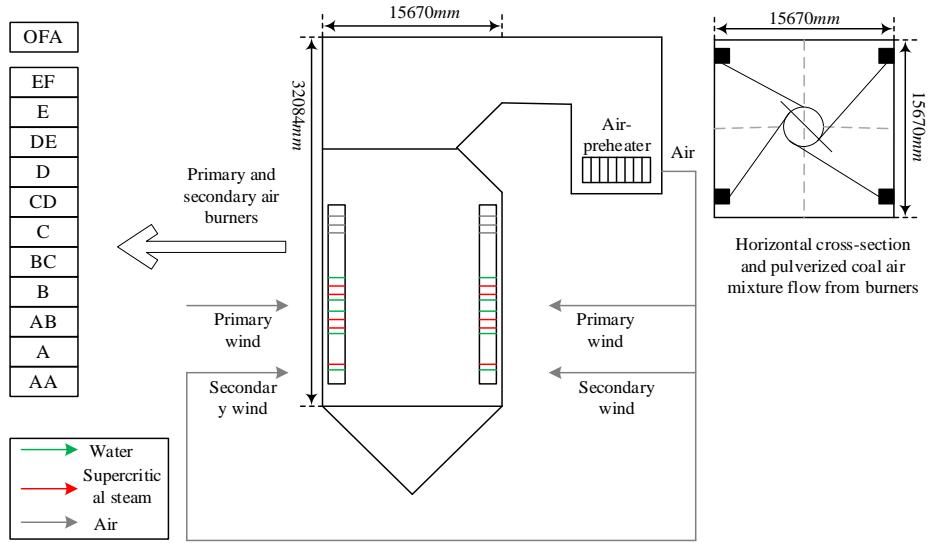

**Fig. 1.** Schematic diagram of the boiler.

## 2.2 Variables preliminary selection based on mechanism analysis of NOx generation.

NOx emission concentration generated in the combustion process include NO (95%) and NO2 (5%) [28,29]. According to the mechanism of different NOx, there are two main types of NOx produced by coal-fired boilers: fuel-type NOx and thermal-type NOx [30].

Thermal-type NOx is the nitrogen oxide produced by the oxidation of N2 in the burning air at high temperatures. The formation reaction of NO is formulated as Eq. (1), also known as the Zeldovich model. The Zeldovich formula is defined as Eq. (2), indicating that the main influencing factors of thermal NOx are temperature and oxygen concentration.

$$\begin{cases} O_2 \rightleftarrows O+O \\ N_2 + O \rightleftarrows NO + N \\ N + OH \rightleftarrows NO + H \end{cases} \quad (1)$$

$$\begin{cases} d[NO]/dt = 2k_1 k_0 [N_2][O_2]1/2 \\ 2k_1 k_0 = k = 3 \times 10^{14} e^{-54200/(RT)} \end{cases} \quad (2)$$

Fuel-type NOx accounts for 65%~85% of the total NOx. According to Hein [31], the generation of fuel NOx is closely related to the pyrolysis product of coal and the oxygen concentration in the flame, where the oxygen concentration and its distribution play a decisive role in the production of NOx. According to the mechanical analysis, preliminarily selected input variables related to NOx emission concentrationare shown in Table 1.

**Table 1**
Primary variables related to NOx.

| Number | Parameter description | Symbol | Unit | Lower limit | Upper limit |
|---|---|---|---|---|---|
| 1 | Secondary airflow (AA) | $S_{AA}$ | m/s | 72.78 | 130.38 |
| 2 | Secondary airflow (AB) | $S_{AB}$ | m/s | 88.64 | 182.50 |
| 3 | Secondary airflow (BC) | $S_{BC}$ | m/s | 101.25 | 102.75 |
| 4 | Secondary airflow (CD) | $S_{CD}$ | m/s | 59.66 | 152.15 |
| 5 | Secondary airflow (DE) | $S_{DE}$ | m/s | 144.10 | 261.59 |
| 6 | Secondary airflow (EF) | $S_{EF}$ | m/s | 138.60 | 234.92 |
| 7 | Primary airflow(A) | $P_A$ | m/s | 41.30 | 57.36 |
| 8 | Primary airflow(B) | $P_B$ | m/s | 41.42 | 57.23 |
| 9 | Primary airflow(C) | $P_C$ | m/s | 41.54 | 57.17 |
| 10 | Primary airflow(D) | $P_D$ | m/s | 41.97 | 56.69 |
| 11 | Primary airflow(E) | $P_E$ | m/s | 42.03 | 56.56 |
| 12 | Primary airflow(F) | $P_F$ | m/s | 42.16 | 56.69 |
| 13 | Coal feed rate($F_1$) | $F_1$ | t/h | 139.07 | 163.38 |
| 14 | Coal feed rate($F_2$) | $F_2$ | t/h | 120.28 | 158.16 |
| 15 | Coal feed rate($F_3$) | $F_3$ | t/h | 127.26 | 162.20 |
| 16 | Coal feed rate($F_4$) | $F_4$ | t/h | 115.15 | 150.79 |
| 17 | Coal feed rate($F_5$) | $F_5$ | t/h | 123.75 | 172.97 |
| 18 | Coal feed rate($F_6$) | $F_6$ | t/h | 115.39 | 150.13 |
| 19 | Total coal rate | $T_C$ | t/h | 232.74 | 382.48 |
| 20 | Main steam pressure | P | Mpa | 21.94 | 32.09 |
| 21 | Main steam temperature | T | ℃ | 579.16 | 612.12 |
| 22 | Oxygen concentration | $O_2$ | % | 1.65 | 5.38 |
| 23 | Total air volume | TV | t/h | 2257.22 | 3789.66 |

## 3. Model development

In order to establish an accurate NOx emission concentration prediction model, a NOx emission concentration prediction algorithm, AE-ELM, is proposed. The algorithm consists of MI-based feature selection, delay time determination, deep learning-based feature extraction, and ELM modeling, which are respectively explained in sections 3.1 to 3.4. The main process of the algorithm is presented in section 3.5.

### 3.1 MI-based feature selection

The variables related to NOx emission concentration are obtained through mechanism analysis (Section 2.2). These variables have nonlinear relationships with NOx concentrations to varying degrees. MI algorithm analyzes the correlation between these variables and NOx emission concentration.

MI measures the correlation between events X and Y based on the principle of information entropy [32]. The degree of correlation between the two random variables can be expressed by the MI as follows [33]:

$$H(X) = -\sum_{i=1}^{n} p(x_i) \log p(x_i) \tag{3}$$

$$I(X;Y) = H(X) + H(Y) - H(X,Y) \tag{4}$$

where $H(X)$ and $H(Y)$ are the information entropy values of $X$ and $Y$, respectively. $X = [x_1, x_2, \cdots, x_n]$, $n$ is the number of samples in dataset $X$, $p(x_i)$ is the probability distribution of $x_i$. $H(X,Y)$ is the joint entropy of $X$ and $Y$.

$D = \{x(k), y(k)\}$ is the data set used for modeling, where $x(k) = [x_1(k), x_2(k), \cdots, x_m(k)]$. $m$ is the dimension of primary variables. $y(k)$ is the NOx emission concentration, and $k$ is the number of samples. Then, the MI is computed as follows:

$$I(x;y) = \sum_{y} \sum_{x \in X} p(x,y) \log \frac{p(x,y)}{p(x)p(y)} \tag{5}$$

where $p(x)$ and $p(y)$ are the probability density functions of x and y, $p(x,y)$ is a joint probability density function of $x$ and $y$. The obtained 23 variables make up the initial input variable set $x(k)$, where $x(k)=[x_1(k), x_2(k),\cdots,x_m(k)]$. $y(k)$ is the NOx emission concentration. In data statistic theory, the correlation coefficient MI between two variables greater than 0.4 indicates that the two variables are moderately or highly correlated. Therefore, 13 features with coefficients greater than 0.6 are selected as input variables for subsequent prediction of NOx concentration. Table 2 presents the ranking of the importance of each feature.

**Table 2**
The analysis of MI and time delay.

| NO | Variables | MI | Delay time(min) |
|---|---|---|---|
| 1 | $F_1$ | 0.76 | 6 |
| 2 | $F_2$ | 0.77 | 6 |
| 3 | $F_3$ | 0.76 | 6 |
| 4 | $S_{AB}$ | 0.87 | 5 |
| 5 | $S_{CD}$ | 0.86 | 4 |
| 6 | $S_{EF}$ | 0.88 | 4 |
| 7 | $P_A$ | 0.81 | 5 |
| 8 | $P_C$ | 0.80 | 5 |
| 9 | P | 0.71 | 3 |
| 10 | T | 0.65 | 5 |
| 11 | $T_C$ | 0.68 | 5 |
| 12 | TV | 0.65 | 4 |
| 13 | $O_2$ | 0.70 | 3 |

## 3.2 Delay time determination

According to practical experience, A time delay exists between the 13 variables (selected in Section 3.1) and NOx emission concentration. To estimate the delay time only from the mechanism analysis will make the certainty cannot be guaranteed. Therefore, in this study, the delay time between each feature and the NOx emission concentration is determined by the MI method. The data set $D=\{x(k), y(k)\}$ is used in modeling, where $x(k)=[x_1(k), x_2(k),\cdots,x_{13}(k)]$, $y(k)$ is NOx emission concentration. A certain time delay exists between $x_i(k)$ and $y(k)$. Thus, the feature values before the time $d_i$ are related to $y(k)$, i.e., $x_i(k-d_i)$ and $y(k)$, where $d_i$ is the delay time between the features $x_i(k)$ and the NOx emission concentration $y(k)$. $x_i(k-d_i)$ is the data of $x_i(k)$ time $d_i$ ago. Then, the input variable matrix is reconstructed as follows:

$$X(k) = [x_1(k), x_2(k), \cdots, x_{13}(k)] \tag{6}$$
$$\tilde{X}(k) = [x_1(k-d_1), x_2(k-d_2), \cdots, x_{13}(k-d_{13})] \tag{7}$$

Through the mechanism analysis of the generation of NOx emission concentration, $d_{max}$, the maximum possible delay time for all input variables, can be determined. According to the actual operation of the boiler selected in the experiment, $d_{max}$ is set to 10 min. The correlation between the reconstructed input data and NOx output is analyzed by the MI algorithm. The results are given in Table 2.

### 3.3 Automatic encoder

In Section 3.2, the set of input features $\mathbf{x}(k) = [x_1(k), x_2(k), \cdots, x_{13}(k)]$ is re-represented as $\tilde{\mathbf{x}}(k) = [\tilde{x}_1(k), \tilde{x}_2(k), \cdots, \tilde{x}_{13}(k)]$, where $\tilde{x}_i(k) = x_i(k-d_i)$, and $d_i$ is the delay time of the input varibale $x_i(k)$.

$[\tilde{x}_1(k), \tilde{x}_2(k), \cdots, \tilde{x}_{13}(k)] = [x_1(k-5), x_2(k-5), \cdots, x_{13}(k-6)]$ (Section 3.2). In this section, the AE is used to reduce the dimensionality of data, where the number of hidden layer nodes is lower than the input layer. The structure of AE is illustrated in Fig. 2.

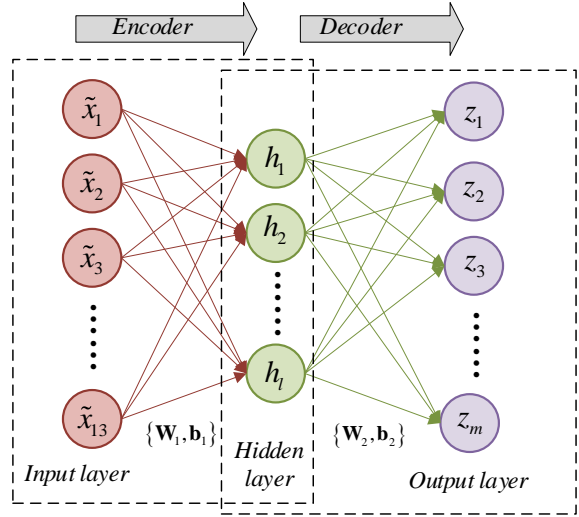

**Fig. 2.** Model structure of AE.

From input layer ($L_1$) to hidden layer ($L_2$), which is the encoding process, $\tilde{\mathbf{x}} = [x_1(k-5), x_2(k-5), \cdots, x_{13}(k-6)]$ is mapped to the intermediate hidden layer $L_2$, formulated as follows:

$$\mathbf{h} = f(\mathbf{W}_1 \tilde{\mathbf{x}} + \mathbf{b}_1) \tag{8}$$

where $f(\cdot)$ is the sigmoid activation function. $\mathbf{W}_1$ and $\mathbf{b}_1$ are the weight matrix and the bias vector of the encoding stage, respectively. $\mathbf{h}$ is the output of the hidden layer. The decoding processing, from the hidden layer ($L_2$) to the output layer ($L_3$), the output layer $L_3$ is computed as follows:

$$\mathbf{z} = f(\mathbf{W}_2 \mathbf{h} + \mathbf{b}_2) \tag{9}$$

where $\mathbf{W}_2$ is the weight matrix between layers $L_2$ and $L_3$, $\mathbf{b}_2$ is the offset vector of $L_3$. The output $\mathbf{z}$ is the reconstruction of input $\mathbf{x}$, which contains useful information in $\mathbf{x}$. The following reconstructed error is used as a loss function:

$$\underset{\theta_1,\theta_2}{\operatorname{argmin}} \|\mathbf{z} - \tilde{\mathbf{x}}\|_2^2 = \operatorname{argmin} \|f(\mathbf{W}_2 f(\mathbf{W}_1 \tilde{\mathbf{x}} + \mathbf{b}_1) + \mathbf{b}_2) - \tilde{\mathbf{x}}\|_2^2 \tag{10}$$

where $\theta_1$ and $\theta_2$ are the parameter set of the encoder and decoder, respectively. $\theta_1 = \{\mathbf{W}_1, \mathbf{b}_1\}$, $\theta_2 = \{\mathbf{W}_2, \mathbf{b}_2\}$. Therefore, the AE extracts the internal nonlinear relationship of the input dataset, mines the important deep feature, and minimizes the reconstructed error. Ten experiments are conducted to determine the number of output neurons $\mathbf{z} = [z_1(k), z_2(k), \cdots z_{10}(k)]$. The MAPE over the ten experiments is plotted in Fig. 3.

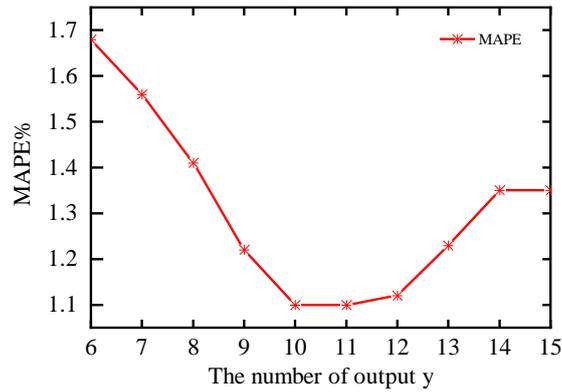

**Fig. 3.** Searching process of the output y.

## *3.4 Modeling of NOx emission concentration with the ELM*

The ELM is a machine learning method that can effectively solve various classification and regression problems. Fig. 4 illustrates the structure of ELM. In the ELM, the input weight matrix $\mathbf{W}$ is randomly generated, and the output weight matrix $\mathbf{\beta}$ is determined analytically, unlike other machine learning models that iteratively adjust all the parameters of the network.

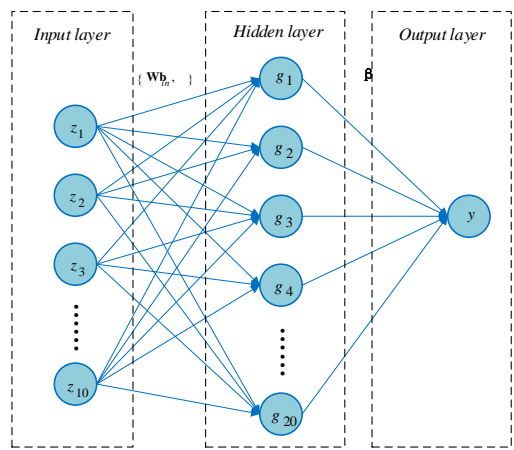

**Fig. 4.** Structure of ELM.

$\mathbf{z}_j = [z_1(k), z_2(k), \cdots, z_{10}(k)]^T \in R^n$ is an $n$-dimensional feature matrix (a matrix with ten variables), and $y_j \in R$ is the target vector (the NOx emission concentration). In order to develop an ELM-based NOx emission concentration model with $N$ samples $(\mathbf{z}_j, y_j)$, the ELM with $K$ hidden neurons can be expressed as:

$$\sum_{i=1}^{K} \beta_i G_i(\mathbf{z}_j) = \sum_{i=1}^{K} \beta_i g(\mathbf{w}_i \cdot \mathbf{z}_j + b_i) = y_j, j = 1, \cdots, N \quad (11)$$

Where $\mathbf{w}_i = [w_{i1}, w_{i2}, \cdots, w_{i20}]$ is the weight vector connecting the input neuron and the $ith$ hidden neuron, $b_i$ is the threshold of the $ith$ hidden neuron, and $\beta_i$ is the weight between the $ith$ hidden neuron and the output neuron. The operator $\mathbf{A} \cdot \mathbf{B}$ in Eq. (11) denotes the inner product of $\mathbf{A}$ and $\mathbf{B}$. $G_i(\mathbf{z}_j)$ is the output of the $ith$ hidden neuron.

The dynamic prediction model (AE-ELM) is established using the preprocessed data. The preprocessing (feature selection and extraction) can reduce the data dimension to eliminate some coupling-related variables, thereby significantly improving the modeling accuracy while reducing the computational complexity. The modeling time of AE-ELM is reduced from 98 secs to 57 secs, reducing 41.8% computational time, as shown in Fig. 5.

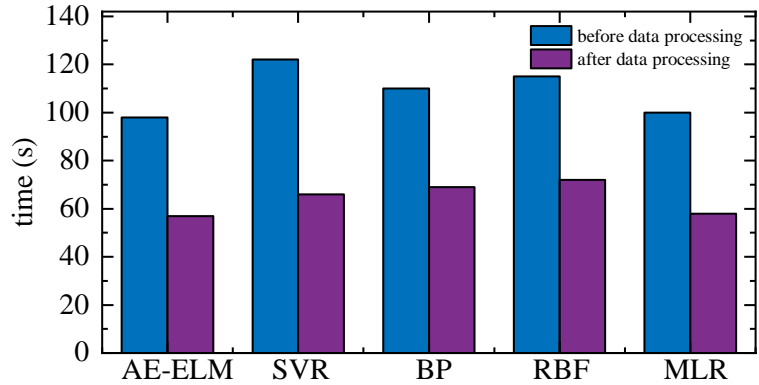

**Fig. 5.** The computational time of different models with/without data preprocessing.

## 3.5 The main algorithm

The overall flowchart of the proposed method is depicted in Fig. 6. The proposed method is conducted in the following steps.

**Step 1:** The historical operation data suitable for model construction is collected from the distributed control system (DCS) that records the historical data of operating parameters in power plants. Then the mechanism analysis is conducted to obtain the initial data set $\mathbf{d}_i = [x_1(k), x_2(k), \cdots, x_{23}(k)]$.

**Step 2:** The MI between each input variable and NOx concentration is computed to select input features, obtaining the data set $\mathbf{x}_i = [x_1(k), x_2(k), \cdots, x_{13}(k)]$.

**Step 3:** The delay time between each input feature and NOx concentration is determined by the MI algorithm, obtaining the data set $\tilde{\mathbf{x}}_i = [x_1(k-3), x_2(k-5), \cdots, x_{13}(k-6)]$.

**Step 4:** The AE-based feature extraction is conducted on the data set $\tilde{\mathbf{x}}_i = [x_1(k-3), x_2(k-5), \cdots, x_{13}(k-6)]$ to obtain a new data set $\mathbf{z}_i = [x_1(k), x_2(k), \cdots, x_{10}(k)]$.

**Step 5:** The ELM prediction model is trained, and the prediction result is computed as $y_j = \sum_{i=1}^{K} \beta_i g(\mathbf{w}_i \cdot \mathbf{z}_j + b_i), j = 1, \cdots, N$.

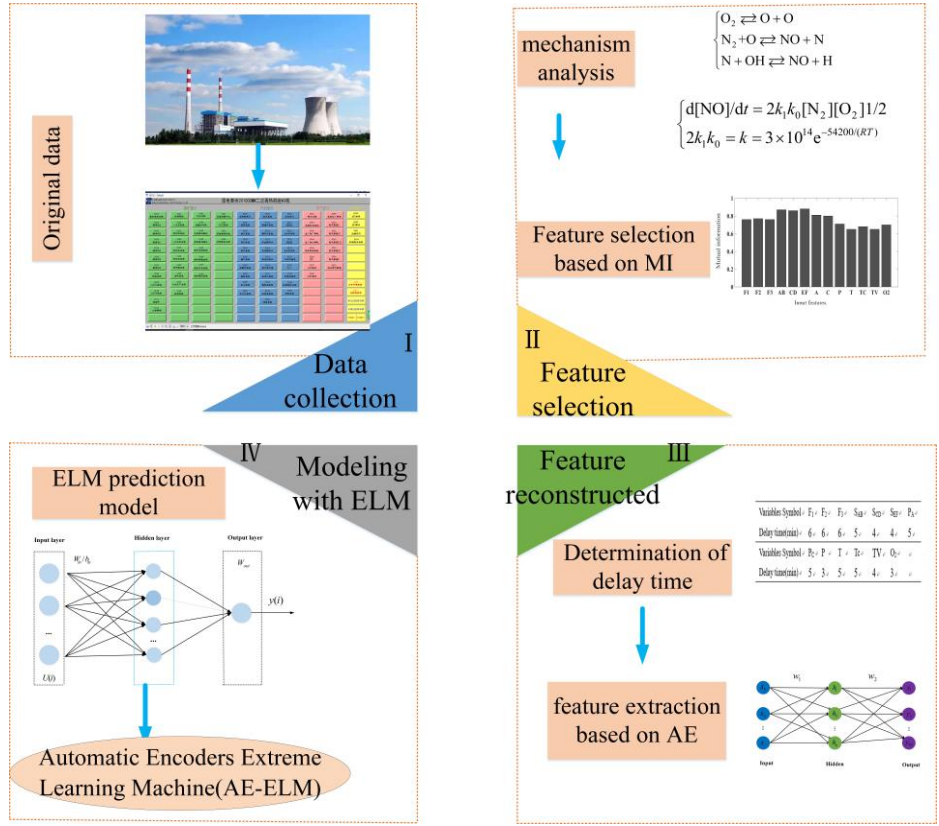

**Fig. 6.** Flowchart of the proposed algorithm.

## 4. Experimental results and discussion

### *4.1 Experimental design*

The effectiveness of the proposed model is validated on the historical data from the DCS of a 1000MW ultra-supercritical boiler in a thermal power plant. Experimental data were collected randomly with a sampling time of 1min. The 500 historical data, selected and processed as described in Section 3, was used to construct the prediction model. Twenty-three variables affecting Nox emission concentration were selected according to the mechanism analysis of NOx generation, ensuring the reasonableness of the initial variables selection and facilitating further selection of variables with a strong nonlinear relationship with NOx emission concentration as model inputs. The effectiveness of the prediction model was verified on the operating data of the power plant. Also, four quarters were used as the validation data to evaluate the universality and generalizability of the prediction model. 500 samples were selected in each quarter; thus, these four data sets span a large period, fully reflecting the operating characteristics of power plants in different periods. For the convenience of subsequent description, the four data sets are represented as D1, D2, D3, and D4, as shown in Table 3. The proposed AE-ELM model was compared with the multiple linear regression (MLR) model, BP model, RBF model, and SVR model [34].

**Table 3**
Experimental dataset.

| Dataset | Train set | Test set | Collection data |
|---------|-----------|----------|-----------------|
| D1 | 400 | 100 | 2016.1.1 |
| D2 | 400 | 100 | 2016.4.1 |
| D3 | 400 | 100 | 2016.9.1 |
| D4 | 400 | 100 | 2016.12.1 |

## 4.2 Performance evaluation criteria

In order to evaluate the prediction performance, three evaluation indexes are used: Mean relative error (MAPE), the normalized average of the squares of the errors (NMSE), and R-Squared ($R^2$). The smaller the values of MAPE and NMSE, the better the prediction performance of the model. The closer the $R^2$ is to 1, the more accurate the prediction performance of the model is. They are mathematically defined as follows:

$$MAPE = \frac{1}{n}\left[\sum_{i=1}^{n}\frac{|y_i - \hat{y}_i|}{\hat{y}_i}\right] \times 100\% \qquad (12)$$

$$NMSE = \frac{1}{n}\sum_{i=1}^{n}\frac{(\hat{y}_i - y_i)^2}{y_i \cdot \hat{y}_i} \qquad (13)$$

$$R^2 = 1 - \frac{\left(\sum_{i=1}^{n}(\hat{y}_i - y_i)^2\right)/n}{\left(\sum_{i=1}^{n}(y_i - \bar{y}_i)^2\right)/n} \qquad (14)$$

where $n$ represents the total number of test samples, $y_i$ is the measured value of NOx emission concentration, and $\hat{y}_i$ is the predicted value of NOx emission concentration.

## 4.3 Experimental results and analysis

### 4.3.1 The effectiveness of time delay analysis

An ablation study was conducted to verify the influence of time delay analysis on the prediction. As summarized in Table 4, the prediction accuracy of all five models has been improved with the time delay analysis. The AE-ELM model decreased by 0.12%, and MLR, BP, RBF, and SVR models all decreased by more than 0.2% in terms of MAPE. As for NMSE, the AE-ELM model decreased by 4.9%, and MLR, BP, RBF, and SVR models are all decreased by more than 4%. Table 4 shows that the proposed model outperforms the other compared models.

**Table 4**
Prediction results of different models (C: Considering time delay; N: No time delay).

| Models | Process | NMSE | MAPE (%) | $R^2$ |
|---|---|---|---|---|
| AE-ELM | N | 0.0241 | 0.8569 | 0.9887 |
|  | C | **0.0229** | **0.8558** | **0.9891** |
| MLR | N | 0.1956 | 2.3402 | 0.8639 |
|  | C | **0.1915** | **2.3359** | **0.8662** |
| BP | N | 0.0721 | 1.5140 | 0.9759 |
|  | C | **0.0634** | **1.5129** | **0.9768** |
| RBF | N | 0.1123 | 1.7230 | 0.9369 |
|  | C | **0.1104** | **1.7244** | **0.9402** |
| SVR | N | 0.1473 | 1.9965 | 0.9276 |
|  | C | **0.1465** | **1.9979** | **0.9282** |

The proposed AE-ELM model achieved the outstanding average relative error: 7.81% better than the MLR model, 5.24% better than the BP model, 5.87% better than the RBF model, and 6.57% better than the SVR model. In terms of the criteria of average absolute error, the proposed AE-ELM model and the other four models differ by 13.38%, 14.5%, 10.19%, and 12.84% after time analysis, respectively. The analysis of the correlation between input and output time delay concluded that the boiler system is a large time-delay system.

### 4.3.2 The effectiveness of feature selection

This section mainly analyzes the influence of the MI algorithm on prediction accuracy and computational efficiency. Table 5 compares the models with and without the feature selection on modeling by MI method. As shown in Table 5, the $R^2$ of the AE-ELM model is increased by 0.92%, the MAPE is reduced by 0.3%, and the NMSE is reduced by 14.5% with the adaptation of the MI method. Further, when the feature selection is applied, the computational time of the AE-ELM model is reduced by 14.3 secs, and the modeling time is saved by 35.6%. The results prove that both the efficiency and precision of the prediction model can be improved by reducing the input dimension and eliminating some of the variables with coupling correlation.

**Table 5**
Ablation study for AE-ELM feature selection.

| Error metrics | AE-ELM | |
|---|---|---|
| | With MI | Without MI |
| $R^2$ | **0.9891** | 0.9800 |
| MAPE (%) | **0.8558** | 0.8592 |
| NMSE | **0.0229** | 0.0268 |
| time (s) | **25.80** | 40.10 |

### 4.3.3 The effectiveness of AE feature extraction

This section discusses the influence of feature extraction with AE. The prediction results modeled by the before-and-after comparison algorithm of AE feature extraction were compared. As shown in Fig. 7, the prediction results of all the algorithms improved with the AE algorithm, indicating the effectiveness of AE feature extraction.

Deep mining of the internal information, achieved by deep network AE from the data improved the extraction ability and consequently improved the prediction accuracy of all the algorithms.

The most accuracy-improved algorithm was the SVR, whose MAPE was reduced by 25% and NMSE reduced by 36% with the extraction. The MAPE of other algorithms all reduced by more than 20% and NMSE by more than 30%.

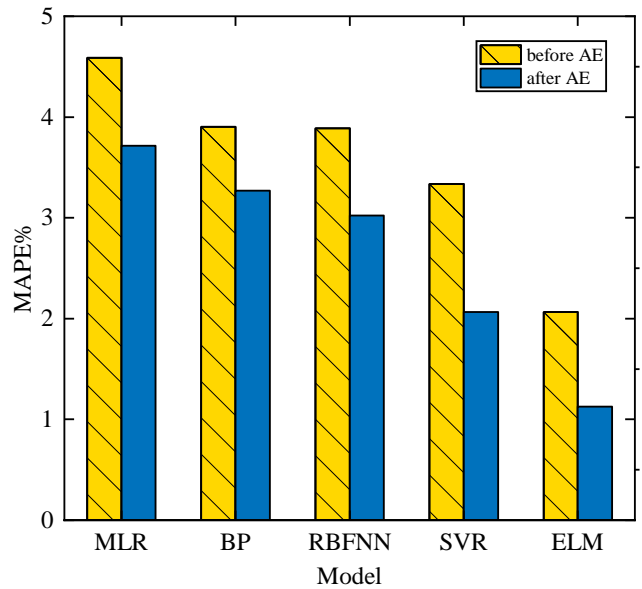

**Fig. 7.** MAPEs with/without AE feature extraction.

*4.3.4 NOx emission concentration prediction model analysis*

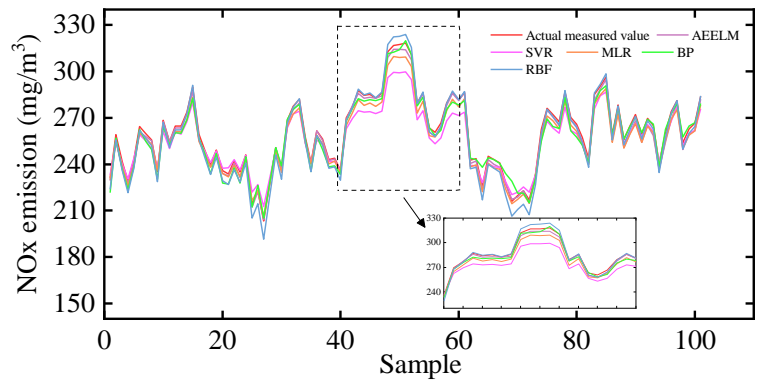

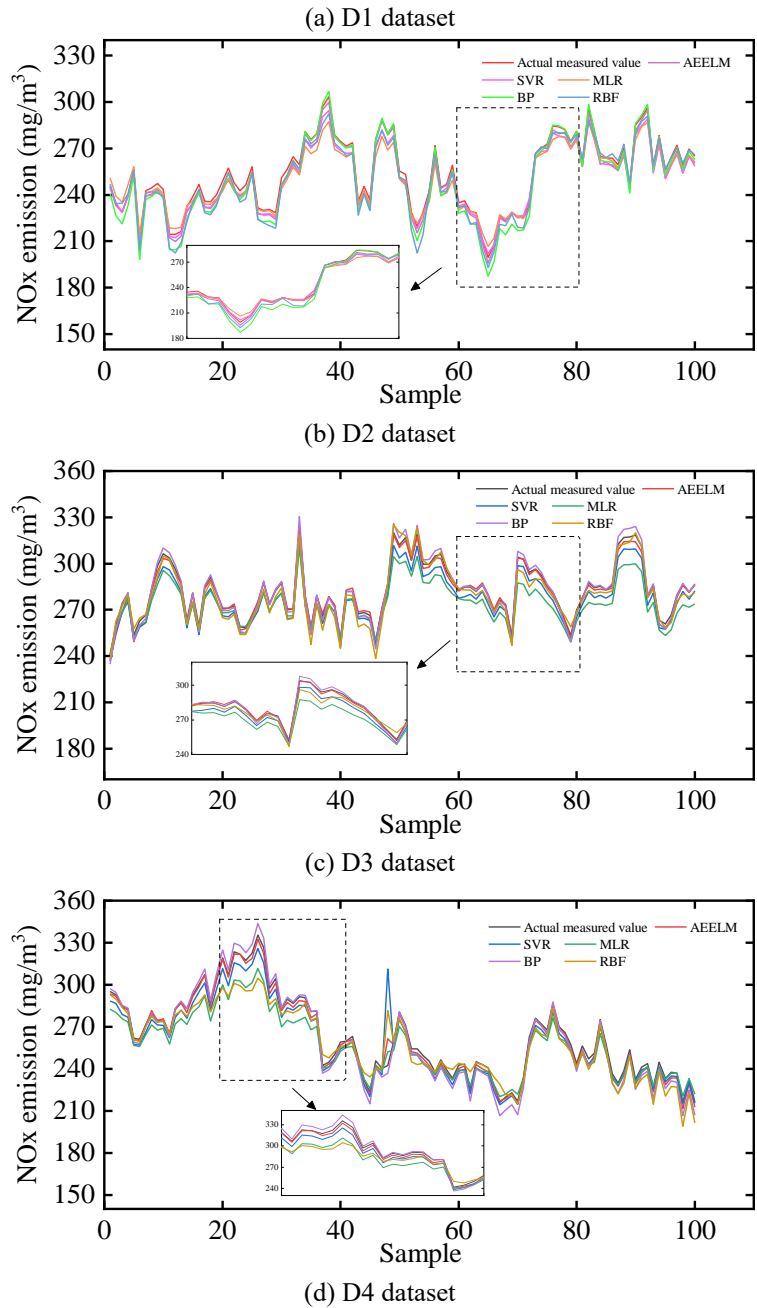

**Fig. 8.** Prediction results of different models.

Fig. 8 compares the prediction accuracies of the compared methods on four validation datasets. The sub-plots depict cropped and magnified partial curves, better comparing the five compared models. It can be clearly seen from Fig. 6 that AE-ELM prediction curves following the direction of the measured data can effectively predict the NOx emission concentration. It indicates that AE-ELM has the best forecasting performance. The MLR algorithm shows the worst prediction performance, where the predicted curves do not reflect the actual trend. This is because the MLR model is not good at dealing with nonlinear problems.

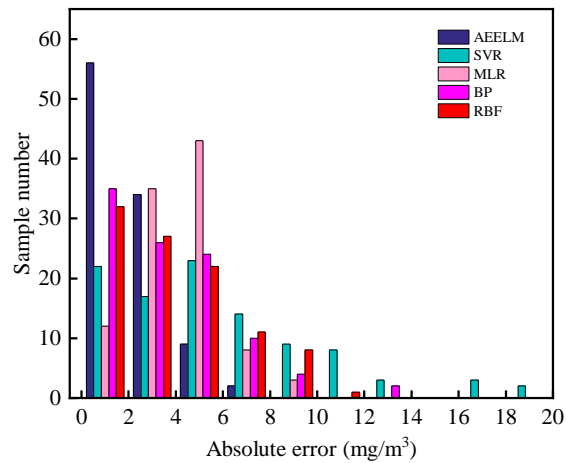

(a) D1 dataset

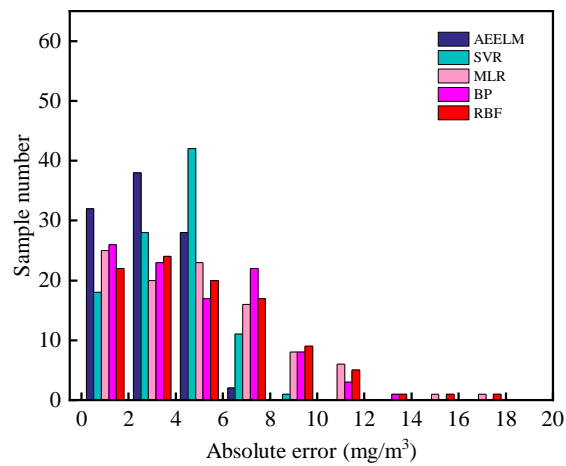

(b) D2 dataset

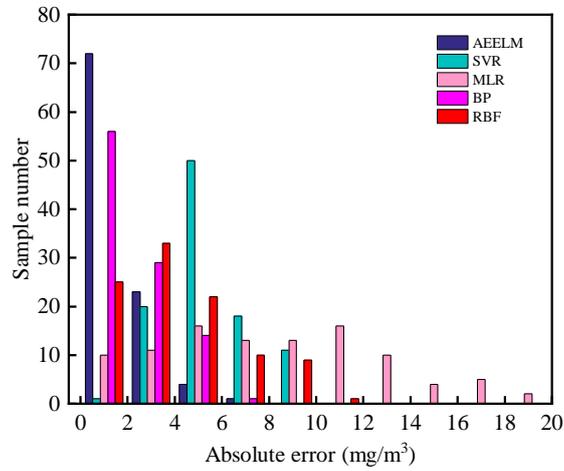

(c) D3 dataset

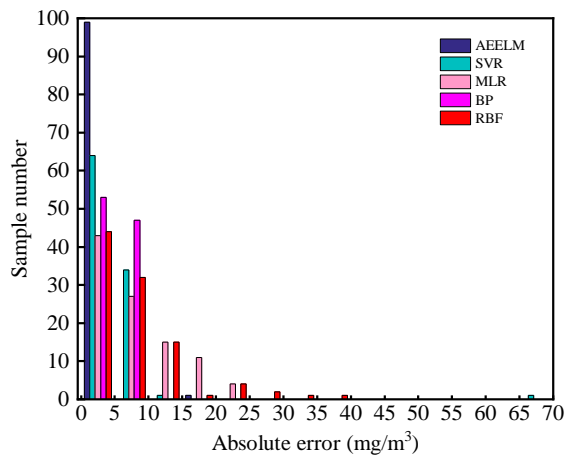

(d) D4 dataset

**Fig. 9.** Absolute errors of different models.

To further validate the performance of the AE-ELM model, the absolute error between the measured values and the predicted values of the five algorithms were compared in Fig. 9. The absolute error of the AE-ELM model was distributed within the minimum interval [0, 10], as shown in Fig. 7. Also, the increasing frequency decreases with the increase of the absolute error. The frequency distribution of MLR, BP, RBF, and SVR models was similar to AE-ELM, but the absolute error frequency of the AE-ELM model decreased rapidly, and there was no distribution in the higher absolute error interval. The AE-ELM model provides the absolute minimum error, i.e., the best prediction accuracy over the other compared methods.

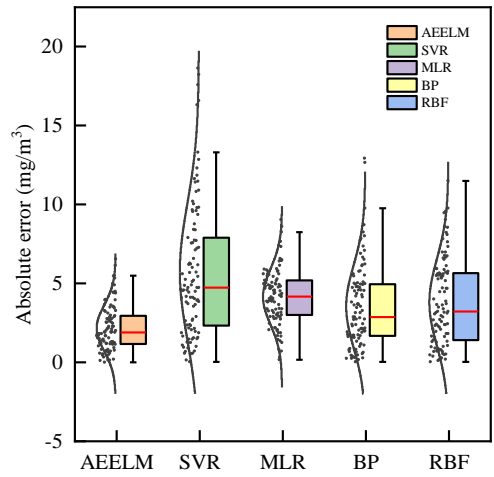

(a) D1 dataset

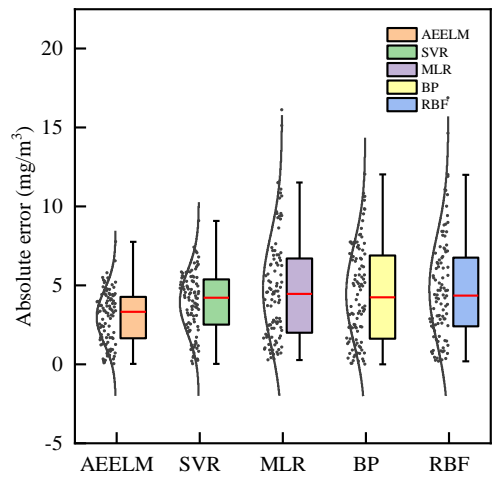

(b) D2 dataset

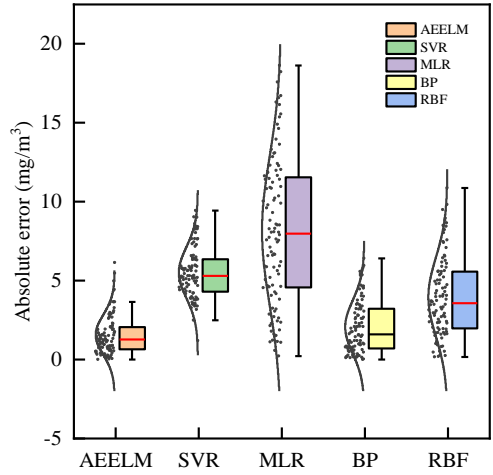

(c) D3 dataset

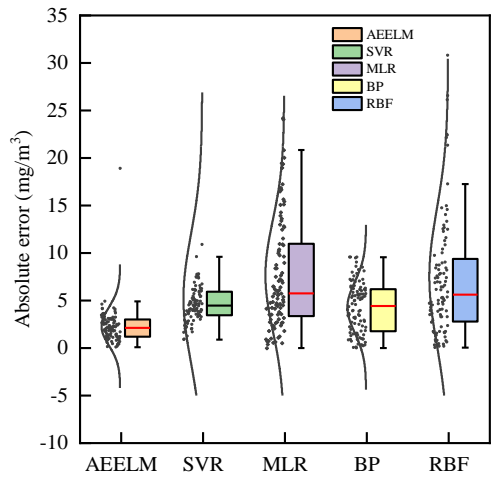

(d) D4 dataset

**Fig. 10.** The boxplot of the absolute errors of different models.

Fig. 10 shows the box plots of the absolute errors between the measured and predicted values for the four datasets, where the red line indicates the median error. The errors of the proposed AE-ELM algorithm vary within a small range, and the lowest errors were obtained over the other four models.

Table 6 summarizes the comparison of the prediction results for five compared models. The analysis shows that the proposed model has the best forecasting performance. In dataset D1, the proposed model achieved the NMSE, MAPE, and $R^2$ of 0.0229, 0.8558%, 0.9891; the BP model obtained the second-best forecasting performance (NMSE, MEPE, and $R^2$ of 0.0634, 1.5129%, 0.9768) compared with other individual models. Thus, the proposed model outperformed the second-best model, BP, by large margins: 63.8% NMSE, 43.43% MAPE, 96.44% $R^2$, respectively.

In dataset D2, the proposed model achieved the NMSE, MAPE, and $R^2$ of 0.0231, 0.8560%, 0.9892; compared with other individual models, the BP model has the second-best forecasting performance (NMSE, MEPE, and $R^2$ of 0.0636, 1.5130%, 0.9766). The proposed model outperformed the second-best model, BP, also on D2 by large margins: NMSE, MAPE, $R^2$ are respectively 63.6%, 43.42%, and 96.44% lower than the BP network. The other two datasets show the same trend, proving the proposed AE-ELM model achieves higher accuracy and better fitting prediction ability.

**Table 6**
Modeling performance of AE-ELM and other models.

| Dataset | Performance criteria | AE-ELM | BP | RBF | SVR | MLR |
|---|---|---|---|---|---|---|
| D1 | *NMSE* | 0.0229 | 0.0634 | 0.1104 | 0.1465 | 0.1915 |
| | *MAPE (%)* | 0.8558 | 1.5129 | 1.7244 | 1.9979 | 2.3359 |
| | *$R^2$* | 0.9891 | 0.9768 | 0.9402 | 0.9282 | 0.8662 |
| D2 | *NMSE* | 0.0231 | 0.0636 | 0.1106 | 0.1468 | 0.1917 |
| | *MAPE (%)* | 0.8560 | 1.5130 | 1.7245 | 1.9980 | 2.3360 |
| | $R^2$ | 0.9892 | 0.9766 | 0.9401 | 0.9281 | 0.8661 |
| D3 | *NMSE* | 0.0228 | 0.0633 | 0.1103 | 0.1464 | 0.1913 |
| | *MAPE (%)* | 0.8556 | 1.5128 | 1.7243 | 1.9978 | 2.3358 |
| | *$R^2$* | 0.9889 | 0.9769 | 0.9403 | 0.9283 | 0.8663 |
| D4 | *NMSE* | 0.0230 | 0.0635 | 0.1105 | 0.1466 | 0.1916 |
| | *MAPE (%)* | 0.8559 | 1.5130 | 1.7245 | 1.9980 | 2.3360 |
| | *$R^2$* | 0.9890 | 0.9767 | 0.9401 | 0.9279 | 0.8661 |

## *4.4 Discussion*

The following conclusions are drawn from the analysis of the experimental results:

(1) A certain time delay exists between the input features and NOx emission concentration. Considering the delay time can improve the modeling accuracy. A boiler is a big time-delay system, and the delay time of different input features depends on the time of entering furnace time and the complexity of internal boiler combustion.

(2) Data preprocessing helps to improve the modeling efficiency and accuracy. The feature selection (Section 3.1) and the feature extraction (Section 3.3) significantly improve the modeling efficiency while reducing the computational time. The feature selection and extraction algorithms can reduce the dimension of variables to eliminate some coupling-related variables, thereby improving the modeling accuracy while reducing the time needed for modeling.

## 5. Conclusions

This paper proposes an advanced hybrid model to improve the precision and efficiency of NOx prediction. First, 23 parameters affecting NOx generation were selected as the initial input variables of the model through mechanism analysis. Then, the MI algorithm is used to obtain the input features, considering the correlation of time delay between input features and NOx emission concentration. Third, with AE as the feature extraction, the final ten features are obtained and used for ELM modeling. The experimental results show that the proposed model achieves performance improvement of 1.01%, 0.026, and 0.98 in terms of MAPE, NMSE, and $R^2$, respectively. It indicates a promising prediction performance for NOx emission concentration.

## Acknowledgments

This work was supported in part by the Jilin Science and Technology Project under Grant 20200401085GX and Grant 20190201095JC.